\documentclass{article} 
\usepackage{nips_adapted,times}
\usepackage{hyperref}
\usepackage{url}
\usepackage{graphicx}
\usepackage[utf8]{inputenc}
\usepackage[english]{babel}
\usepackage{graphicx}
\usepackage{amsmath}

\title{Variational Inference via Transformations on Distributions\footnote{\href{https://github.com/jaivardhankapoor/vaenf}{Github Repository for Code in Tensorflow}}}

\author{
Shibhansh Dohare \\
IIT Kanpur\\
\texttt{14644}\\
\texttt{sdohare@iitk.ac.in}\\
\And
 Siddhartha Saxena \\
 IIT Kanpur\\
\texttt{150719}\\
\texttt{siddsax@iitk.ac.in}\\
\And
 Jaivardhan Kapoor\\
IIT Kanpur\\
\texttt{150300}\\
\texttt{jkapoor@iitk.ac.in}\\
}

\newcommand{\Lagr}{\mathcal{L}}

\nipsfinalcopy

\begin{document}

\maketitle

\begin{abstract}
    Variational inference methods often focus upon the problem of efficient model optimization, with little emphasis on the choice of the approximating posterior. In this paper we review and implement the various methods that enable us to develop a rich family of approximating posteriors. We show that one particular method employing transformations on distributions, results in developing very rich and complex posterior approximation. We analyze its performance on the MNIST dataset by implementing with a Variational Autoencoder, and demonstrate its effectiveness in learning better posterior distributions.
\end{abstract}

\section{Introduction}
Posterior computation is one of the central problems in Bayesian Machine Learning, in cases when we have conjugate priors and likelihood models, posterior can be calculated exactly in closed form. Although most of the time, our models are much more complicated and hence we have to deal with non-conjugacy. This inspired researchers to approximate the posterior via some heuristics. Two of the most popular heuristics to approximate the posterior are Variational Inference and Markov Chain Monte-Carlo sampling. 

Variational Inference and MCMC sampling come from two different communities. MCMC has been primarily been studied by Statisticians whereas, Variational Inference has been studied extensively by the Machine Learning community. This has been primarily been due to the fact that VI is a much faster way to approximate the posterior whereas it lacks the guarantee of convergence at an infinite number of iterations.

In this work, we first take a brief survey of some popular Variation Inference techniques and then study and replicate the results of one of the most radical approaches in recent times in VI. 

\section{Related Approaches : A Brief Survey of Variational Inference Techniques}

Variation Inference uses a tight lower bound on the model evidence or ELBO. It tries to maximize it with respect to a variational probability distribution that approximates the posterior.
The core objective of variational inference relies on finding the optimal variational distribution or \textit{\textbf{q(z)}} with some approximations on it that reduce the complexity of the ELBO. 
These approximations inherently limit the performance of the algorithm, hence it forms a possible source of improvement that many researchers have explored over the years. 

\subsection{Mean Field Assumption}

In their introductory papers on variational inference, Jordan \textit{et al.}\cite{Jordan:1999:IVM:339248.339252} introduced the mean field assumption in order to simplify the form of \textbf{ELBO}. The mean field assumption assumes that the variational distribution of all the variables\\ are independent of each other.
\begin{center} $q(\textbf{Z}) = \prod_{i=1}^{i=N}q(z_i)$ \end{center}

This assumption leads to the reduction of ELBO to the summation of ELBOs for each latent variable. This assumption has a major drawback. It does not capture the dependencies between underlying posterior distributions of the latent variables effectively, zero-forcing and underestimates the variations in the data.

Some proposals have been made for richer posterior approximations especially using structured approximations where the approximations try to capture the dependencies in the data. The next sections introduce some schemes to soften this assumption. 

\subsection{Combining the goods of two worlds VI and sampling}

Variational Inference uses an optimization approach to getting the approximate posterior, and this involves finding the gradient of the parameters. The ELBO is the function that is optimized and it involves finding expectations. As expectations can be hard to compute directly, a different approach is to find the expected value of the gradient via sampling from the variational distribution. This significantly simplifies finding expectation as the variational distributions are generally simpler distributions than the actual posterior (as well as known).

\subsection{Hierarchical Variational Models}
\begin{figure}[h]
\centering
\includegraphics[width=0.5\textwidth]{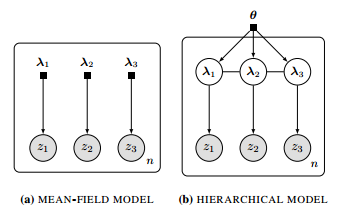}
\caption{An illustration of Hierarchical VI [Figure take from \cite{ICML2016}}
\end{figure}
    A novel approach to link together the distributions of latent variables is to introduce a prior over their parameters. \cite{ICML2016} As illustrated in the figure, there is an underlying parameter above the individual distributions. This makes them dependent on each other and hence relaxes the assumption to some extent. Although this leads to richer models, the expression of the ELBO becomes much more complicated and hence the gradient is calculated through sampling from the variational distribution. 

\subsection{Variational Boosting method}
    
    Boosting is a standard machine learning algorithm in which we train several models sequentially, where at each turn we increase the weights of the misclassified data points Finally all the weak models are combined to produce a strong classifier.
    
    Inspired by this, Miller \textit{et al.} \cite{2016arXiv161106585M} introduced Variational Boosting. \cite{2016arXiv1611.06585H}
The approach is to combine several $q^cs$, each using the mean field assumption and then produce richer distributions using a convex combination of the $q^cs$.  Hence we have

\begin{center} $q^{(C)}(x;\lambda)) = \sum_{c=1}^{C}\rho_c*q_c(x;\lambda)$ \end{center}
    
    Here at each step, two things need to be learned. One is the $q^{(C)}$ and other is $\rho^{(C)}$. Here  $q^{(C)}$  can be calculated as we normally do in VI with mean field assumption, with just one modification that is there is one more latent variable that is the mixing weight for that distribution. Thus it leads to possibly a multi-modal distribution as we have the final posterior a convex sum of several variational distributions.

\subsection{Streaming Variational Bayes}

     A useful modification of the Stochastic VI algorithm is demonstrated in the Streaming Variational Bayes approach\cite{NIPS2013}. In this, streaming distributed and asynchronous computation of the Bayesian posterior is made possible through parallelizing the process. To illustrate it by an example. The work is first divided in K Jobs i.e. we divide the data into K batches and coordinate between these workers through a master, the master holds its own distributions over the latent variables as $\theta_M$. The jobs are run in parallel on K processors. Each processor is given the distribution of latent variables that the master holds at that time and the data.
     As one of the process finishes, the parameters are updated to the global parameters using the following equation. 
     
     \begin{center} $\theta_M' = \theta_i - \theta_M$ \end{center}
     
     Where $\theta_M'$ is the new posterior distribution for master and $\theta_i$ is the posterior distribution learned by the $i_{th}$ worker which was given the prior over parameters as $\theta_M$. 
     
     It can inherently be used as an on-line algorithm as the data size need not be specified, and after every iteration, the posterior approximation is available for us to utilize immediately. The use of exponential families with advantageous properties also aids the process for convenient parameter updates.

These were some of the recent changes brought to VI, the rest of our work focuses on the paper "Variational Inference Normalizing Flows" \cite{2015arXiv150505770J}. It uses flows to transform a simpler distribution though several functions to generate richer posteriors. The technique that is radically different from the above mentioned one and it has gained a lot of attention recently.

The paper's main aim is to demonstrate that instead of just focusing on better optimization methods and expectation computation methods for posterior approximation, we can also choose a richer family of approximate posteriors. This richer family must be large enough that the optimal distribution possibly lies within.

\section{Our Approach}

As mentioned previously, we employ normalizing flows to increase the complexity of the distribution to capture better dependencies in the data. The subsection below describes the theoretical aspects of normalizing flows.

\subsection{A Primer On Normalizing Flows}
An ideal family of variational posteriors $q_{\phi}(z)$ is one that is flexible enough to possibly contain the true posterior. A normalizing flow is a. sequence of \textit{invertible} transformation applied to a probability distribution such that the resulting distribution is also a probability distribution(hence the term \textit{normalising}. The probability densities essentially "flow" through the sequence of transformation on the random variables themselves, and thus a new distribution is obtained.

Normalizing flows may be categorized as finite or infinitesimal. In this paper, we only explore the finite flows. The flow length $K$, defines the number of transformations applied on the base distribution. Consider an invertible function $f \colon {R}^D \to {R}^D$, with inverse $f^{-1} = g$. On appliying f on the random variable $\textbf{z}$ with distribution $p(\mathbf{z})$, we obtain the distribution of $\mathbf{z'} = f(\mathbf{z})$ as 

$$p(\mathbf{z'}) = p(\mathbf{z})\left| det \frac{\partial f^{-1}}{\partial \mathbf{z'}}\right| = p(\mathbf{z})\left| det \frac{\partial f}{\partial \mathbf{z}}\right|^{-1}$$

This equation follows from the result of transformations of probability distributions using Jacobian, and the invertibility of the transforming function.\\
In this way, we can apply a sequence of $K$ transformations upon $\mathbf{z}_0$ with distribution $p(\mathbf{z}_0)$, to obtain a new random variable $\mathbf{z}_K$, with the probability distribution $p(\mathbf{z}_K)$. The expressions for the two are written as 
$$\mathbf{z}_K = f_{K} \circ f_{K-1} \circ \ldots \circ f_1(\mathbf{z}_0)$$
$$p(\mathbf{z}_K) = p(\mathbf{z}_0) \prod_{k=1}^{K} \left| det \frac{\partial f_k}{\partial \mathbf{z}_{k-1}}\right|^{-1}$$

We focus on a specific family of transformations, the invertible linear-time planar flows of the form

$$f(\mathbf{z}) = \mathbf{z} + \mathbf{u}h(\mathbf{w}^T + \mathbf{b})$$

where $h(.)$ is a smooth element-wise non-linearity with derivative $h'(.)$. We can compute the logdet-Jacobian of this trabsformation in $O(D)$ time as follows:

$$\left| det\frac{\partial f}{\partial \mathbf{z}}\right| = \left| det(\mathbf{I} + \mathbf{u}^T\psi (\mathbf{z}))\right|$$
where
$$\psi(\mathbf{z}) = h'(\mathbf{w}^T\mathbf{z} + b)\mathbf{w}$$

Using the above equation, we can determine the log probability after $K$ flows:

$$log~q(\mathbf{z}_K) = log~q(\mathbf{z}_0) - \sum\limits_{k=1}^{K} log\left| det(\mathbf{I} + \mathbf{u}_{k}^{T}\psi_{k} (\mathbf{z}_{k-1}))\right|$$

We can interpret theses flows as expanding/contracting the density perpendicular to the hyperplane $\mathbf{q}^T\mathbf{z} + b = 0$, hence the name planar flows. When the nonlinearity used is $h(x) = tan(x)$, the invertibility of the transformations is ensured by changing the parameters slightly, as demonstrated in the Appendix of \cite{2015arXiv150505770J}. This specific family of transformations is used because it allows for low-cost computation of the determinant.

Consider a model in which we have to infer the latent variables $\mathbf{z}$ given data $\mathbf{x}$. The $ELBO$ of the model, approximated with the posterior $q_{\phi}(\mathbf{z}|\mathbf{x})$, reduces to
$$\Lagr = {E}_{q_{\phi}(z|x)}[log~q_{\phi}(\mathbf{z}|\mathbf{x}) - log~p(\mathbf{z},\mathbf{x})] = {E}_{q_{0}(z_0)}[log~q_{K}(\mathbf{z}_K) - log~p(\mathbf{z}_K,\mathbf{x})] $$ $$= {E}_{q_{0}(z_0)}[log~q_0(\mathbf{z}_0)] - {E}_{q_{0}(z_0)}[log~p(\mathbf{z}_K,\mathbf{x})] - {E}_{q_{0}(z_0)}\left[ \sum\limits_{k =1}^{K}log~\left|1 + \mathbf{u}_k^T \psi_{k}(\mathbf{z}_{k-1})\right|\right]$$

The change of variable with respect to which the expectation is taken is justified by the equation
$${E}_{q_{K}}[g(\mathbf{z})] = {E}_{q_{0}}[g(f_K \circ f_{K-1} \circ \ldots \circ f_{1}(\mathbf{z}))]$$
Notice that the extra term in the $ELBO$ consists of the parameters of the transforming functions, and maximizing the $ELBO$ will optimize these parameters accordingly. 

\subsection{Introduction to Variational Autoencoder (VAE)}

In recent years Variational Autoencoders\cite{2013arXiv1312.6114K} have shown promising results in unsupervised learning to generate data especially in tasks like image generation on MNIST\cite{lecun1998mnist} and CIFAR\cite{Krizhevsky09} data-sets. In generative modeling, we deal with modeling $P(X)$ defined on data points $X$ which typically are images. VAE makes very little assumptions about the data and is very high capacity models. 

Variational auto-encoders assume we have a density distribution $P(z)$ defined over some latent variable z and a deterministic function of the form $f(z|\theta)$ which maps z to space $X$. Our aim is to now optimize $\theta$ so that when we sample z and evaluate $f(z|\theta)$ it's highly likely that $f(z,\theta)$ is in our data-set or is close to some point in the data-set. 

In VAEs the choice of the distribution $p(z)$ is generally a Gaussian, this implies that $P(X|z,\theta) = N(X|f(z,\theta),\sigma^2*I)$. The problem now arise is to choose $P(z)$ such that it captures the latent information. VAEs avoid this information by moving this task to learn latent information from the distribution to the function $f(z,\theta)$. So, we assume that z is coming from a very simple distribution namely $\mathcal{N}(\mathbf{0},\mathbf{I})$. Now, if we have sufficiently powerful function approximators we can learn the mapping from our independently sampled latent variables z to the data points X. This is where Multi-Layer Perceptrons comes to aid as they have shown to be extremely powerful as function approximators.

The next task is to sample those values $z$ that are likely to have produced $x$. This means we need a new function $Q(z|X)$ which can give $z$ that can produce $x$ from these sampled values of z. This helps in the computation of ${E}_{z \sim Q}[p(x|z)]$. The next problem is, how does z sampled from some distribution $q(z|x)$ helps in optimizing $p(x)$. For this we need the relation between ${E}_{z\sim Q} P(X|z)$ and $P(z)$. This relationship constitutes the theoretical working of VAEs, and the final equation\cite{2016arXiv160605908D} turns out to be - 

$$ log P(X) - D[Q(z|X)|P(z|X)] = {E}_{z\sim Q}[log P(X|z)] - D[Q(z)|P(z)]$$

This equation is the core of the VAE. The left side of the equation contains the term $p(x)$ that we're trying to maximize plus an extra term that's trying to make Q produce $z$s that can produce $x$. The first term on the left hand side is like an \textit{encoder} where $Q$ is encoding $x$s into $z$ and $p$ is \textit{decoding} it to reconstruct $x$. The right side of this equation can be directly optimized using stochastic backpropagation\cite{2014arXiv1401.4082J}. Consequently, we will be approximately minimizing our target $p(x)$. Note that second term on the right side of the equation is positive thus the left-hand side acts as a lower bound to $p(x)$. 

\subsection{Implementing Normalizing Flows in VAE}

In this part, we derive the $ELBO$ for the variational autoencoder with normalizing flows, and also describe its architecture. The prior of the VAE, is taken as $p(\mathbf{x}) = \mathcal{N}(\mathbf{0}, \mathbf{I})$, and the encoder is denoted by $q_{\phi}(\mathbf{z})$ with free parameters $\phi$ as the weights of the encoder neural network. Similarly, the decoder is denoted by $p_{\theta}(\mathbf{x}|\mathbf{z})$, with the free parameters $\theta$ as the weights of the decoder neural network. The starting distribution for the application of flows is taken as the reparametrized form of the output of the encoder. After applying flows, we sample from the final distribution $q_K$ and pass it through the decoder to obtain the reconstructed output. A schematic of the architecture is shown in the following figure.

\begin{figure}[h]
\includegraphics[width=10cm]{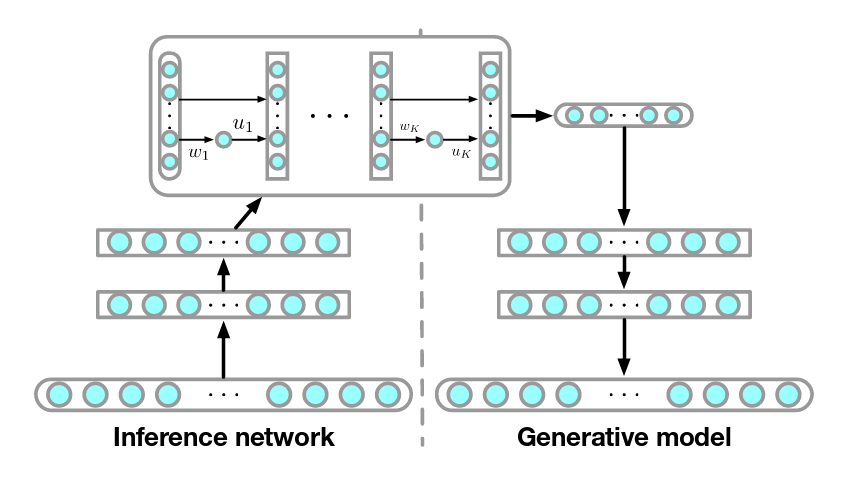}
\centering
\caption{The architecture of the inference network. Borrowed from \cite{2015arXiv150505770J}}
\end{figure}

The $ELBO$ is calculated as
$$\Lagr = {E}_{q_{0}(z_0)}[log~q_0(\mathbf{z}_0)] - {E}_{q_{0}(z_0)}[log~p(\mathbf{z}_K,\mathbf{x})] - {E}_{q_{0}(z_0)}\left[ \sum\limits_{k =1}^{K}log~\left|1 + \mathbf{u}_k^T \psi_{k}(\mathbf{z}_{k-1})\right|\right]$$
which separates into the sum of threee terms - (a) the square error between the input of the encoder and the reconstructed output of the decoder, (b) the $KL$-divergence between the output normal distribution of the encoder and the prior $p(\mathbf{x}) = \mathcal{N}(\mathbf{0}$, and finally the sum of \texttt{logdet-Jacobian} for the extra terms introduced in the ELBO due to the flows (which can me easily represented in terms of the derivative of the nonlinearity used).

We implemented the model using TensorFlow, using the MNIST\cite{lecun1998mnist} dataset for training and testing. The encoder and decoder each had 4 hidden layers with dense connectivity, and the hidden layer dimension was set to 10, with the input and output layer dimension set to 128. The latent dimension was set to 2 for easy visualization of the probability distribution inferred from the dataset. The flow length $K$ was set to 4 and the linearity used was $tanh$. The learning was carried for 500,000 iterations with batch size 1, and the loss optimized using the Adam optimizer\cite{2014arXiv1412.6980K} with a learning rate of 0.002.

\section{Results}
We used a basic VAE architecture to compare our results with respect to that one we observed much more richness in the final distribution of the latent. As we can see in the diagram below, we got a multimodal distribution $p(z|x)$ for the MNIST dataset when we ran with Normalizing flows whereas it was just a normal distribution using a standard VAE. 
\begin{figure}[h]
\centering
\includegraphics[width=0.9\textwidth]{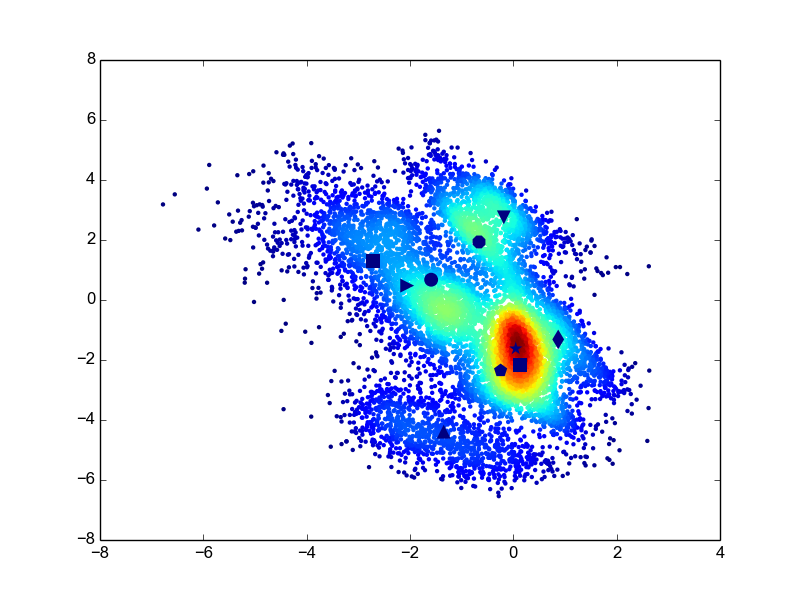}
\includegraphics[width=0.9\textwidth]{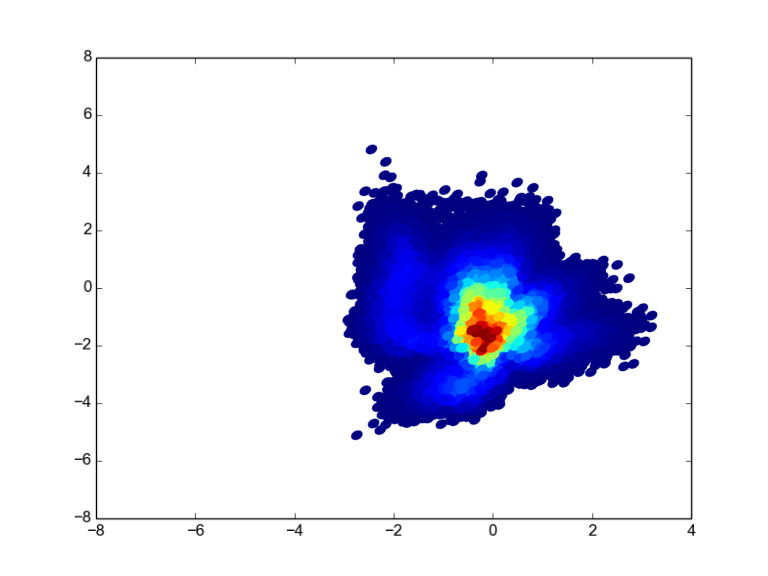}
\caption{Plotting the latent variable value for Normalizing Flows (upper plot) and Standard VAE (lower plot). The shapes in the upper plot corresponds the mean of digits as follows- 0: triangle down, 1: triangle up, 2: diamond, 3: star, 4: circle, 5: square, 6: octagon, 7: square, 8: pentagon, 9: triangle right }
\end{figure}

\begin{figure}
    \centering
    \includegraphics[width=0.9\textwidth]{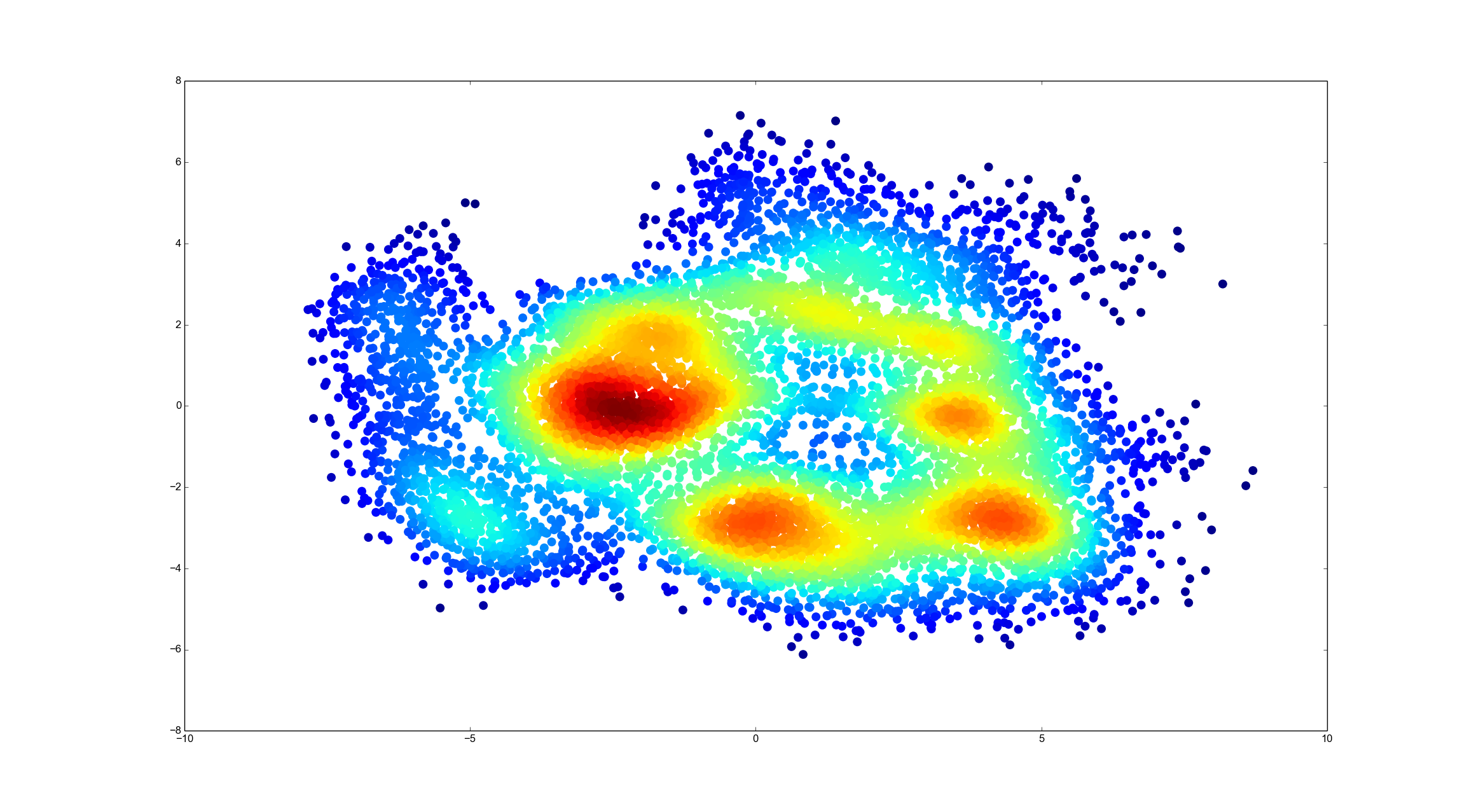}
    \caption{Multimodal nature of the latent variable's distribution }
    \label{fig:my_label}
\end{figure}

To plot the figure, what we did is that we trained the VAE with MNIST dataset and then stored the values of the latent vectors, sampled from the distribution $P(Z|X)$ which is equivalent to sample from $\mathcal{N}(\mathbf{0},\mathbf{I})$, apply the transformation $\mu+\epsilon*\sigma$ and then all the flows over it. 

This hinted that the multimodal distribution represents the different classes and it proved to be the case as well, as shown in the diagram. Thus this shows that Normalizing flows are indeed able to learn much more complicated distributions than what the mean field approximation or one of the above-surveyed methods could do.

\section{Discussion and Future Work}
The results of the experiment hint that the normalizing flows work in a manner similarly to structured VAEs\cite{2016arXiv160208734S}, by stacking many probability distributions on one another for greater correlations between the features and classes of the data.
In the previous sections, we saw that transformations on distributions help aid the inference procedure by helping us explore a larger solution space for the approximate posterior. Although we had run the experiment with only 2 latent dimensions, it gave us significant deviation from standard unimodal distributions obtained in vanilla VAE implementations. Consequently, it is safe to say that increasing the dimensionality will certainly result in more deviation from the prior-type shape of the final distribution.
Variational methods have evolved greatly since their inception, and with the advent of computing technologies like probabilistic programming and parallel processing, the full power of these methods are to be greatly enhanced if used in conjunction with deep learning methods. This remains to serve as a promising avenue for future research. Flexible choices for distributions and structured VAE like richness in the approximate posterior family allow us to explore parallely the fields of optimization and modelling.

\medskip

\end{document}